\documentclass[11pt,a4paper]{article}

\usepackage[margin=1in]{geometry}
\usepackage{graphicx,booktabs,makecell,multirow}
\usepackage{amsmath,amssymb}
\usepackage{siunitx}
\usepackage{hyperref}
\usepackage{etoolbox}
\usepackage{caption}
\usepackage{subcaption}
\usepackage{microtype}

\usepackage{authblk}
\usepackage{orcidlink}

\usepackage[backend=biber,style=numeric,sorting=none]{biblatex}
\usepackage[integrals]{wasysym}
\providecommand{\iq}[1]{``#1''}

\makeatletter
\newif\if@didtitle \@didtitlefalse
\providecommand{\chapter}[1]{%
  \if@didtitle
    \section{#1}
  \else
    \title{#1}%
       
    \author[a,b]{Arianna Miola \orcidlink{0000-0001-6695-8877}}
    \author[c]{Bruno Spaccavento \orcidlink{0009-0004-0379-7518}}
    \author[d]{Lorenzo Silotto \orcidlink{0009-0009-3965-3248}}
    \author[d,e]{Marco Bianchetti \orcidlink{0000-0003-3750-087X}}    
    \author[c]{Luca Cagliero \orcidlink{0000-0002-7185-5247}}
    
    \affil[a]{Intesa Sanpaolo Innovation Center, Corso Inghilterra 3, 10138, Torino, Italy.
        \href{mailto:arianna.miola@intesasanpaolo.com}{\texttt{arianna.miola@intesasanpaolo.com}}}
    \affil[b]{Università degli Studi di Milano-Bicocca, Milan, Italy}
    \affil[c]{Politecnico di Torino, Corso Duca degli Abruzzi 24, 10129, Turin, Italy.
        \href{mailto:luca.cagliero@polito.it}{\texttt{luca.cagliero@polito.it}},
        \href{mailto:bruno.spaccavento@studenti.polito.it}{\texttt{bruno.spaccavento@studenti.polito.it}}}
    \affil[d]{IMI CIB, Market \& Counterparty Risk Management, Intesa Sanpaolo, Milan, Italy.
        \href{mailto:marco.bianchetti@intesasanpaolo.com}{\texttt{marco.bianchetti@intesasanpaolo.com}},
        \href{mailto:lorenzo.silotto@intesasanpaolo.com}{\texttt{lorenzo.silotto@intesasanpaolo.com}}}
    \affil[e]{Department of Statistical Sciences ``Paolo Fortunati'', University of Bologna, Italy}

    \date{\today}
    \date{}
    
    \maketitle
    \@didtitletrue
  \fi
}
\makeatother

\begin{document}

\chapter{Enhancing Financial Question Answering: \\ 
A Novel Benchmark Dataset of Banks' financial statements
}

\begin{abstract}
    The comparative analysis of banks' financial statements poses significant challenges for automated question answering
    system due to their complexity, substantial length, technical language, and inhomogeneity of both textual and numerical content across different jurisdictions and institutions.
     We introduce FinRAG-QA\footnote{https://github.com/Lsilotto/FinRAG-QA}, a novel benchmark dataset for financial question answering, which comprises 999 practitioner-curated questions on 10 standardised indicators, grounded in 209 annual and Pillar 3 reports from 24 major European and U.S. banks spanning 2019-2023. Unlike prior financial QA benchmarks, which centre on U.S. filings and single-institution analysis, FinRAG-QA targets cross-institutional retrieval over documents averaging 198k words, longer than any existing financial QA resource.
    On this benchmark we evaluate a multi-stage RAG pipeline and isolate the contribution of each component. Contextual chunk enrichment combined with a retrieval-optimised embedding model raises NDCG@10 from 0.322 to 0.710; conditional on the ground truth being retrieved, a reasoning-optimised generator raises answer accuracy from 44.6\% to 79.0\% (+34.4 percentage points), at roughly 20× the generation latency. We further show that cross-encoder reranking degrades retrieval when the first-stage ranking is already strong, and that a single top-ranked chunk outperforms larger contexts at generation time. Experiments were run in late 2024–early 2025 with the models available at that time. 
\end{abstract}
 
\noindent\textbf{Keywords:} Document Question Answering, Large Language Models, Retrieval-Augmented Generation, Financial NLP, Benchmark Dataset, Financial Statements, Pillar 3.



\newpage
\tableofcontents
\newpage

\section{Introduction}

In the financial sector, institutions such as banks are required to periodically publish detailed documents illustrating their financial position, capital levels, risk exposure, and liquidity availability. These documents include financial reports prepared according to international accounting standards (IFRS~\cite{ifrs2024navigator} or GAAP~\cite{fasb2024codification}) and Pillar 3 reports prepared according to Basel Committee standards~\cite{basel2023framework}. Although essential for transparency and regulatory oversight, the analysis of these reports presents significant challenges due to their considerable length (often exceeding hundreds of pages), complex technical and legal structure, and inhomogeneity of data representation between different jurisdictions and institutions.

The extraction of key indicators such as capital ratios (e.g., CET1), liquidity reserves, or asset and liability classifications requires substantial time, specialist skills, and cognitive effort from analysts. 
This process is particularly difficult to scale in large-scale comparative contexts, highlighting the need for automated analysis solutions.

Recent advances in Large Language Models (LLMs) have demonstrated remarkable capabilities across various natural language processing tasks. Their application to financial disclosures, however, is far from immediate: corpora of this scale preclude direct in-context processing, and the target is typically a single figure buried in hundreds of pages. LLMs are further limited by hallucinations—generating plausible but factually incorrect responses—and by knowledge cutoffs that prevent access to recent financial data. To address these limitations, Retrieval-Augmented Generation (RAG) frameworks have emerged as a promising solution, combining external knowledge retrieval with generative capabilities.



This paper addresses the critical need for effective financial question answering systems by making the following contributions:

\begin{enumerate}
\item We introduce a novel benchmark dataset containing 999 questions related to key financial indicators across 24 major international banks, spanning 2019-2023.
\item We present a comprehensive RAG architecture specifically optimized for financial document analysis, incorporating contextual chunk enrichment, hybrid retrieval strategies, and advanced reranking techniques.
\item We provide extensive experimental evaluation comparing multiple retrieval approaches, embedding models, and generation strategies, establishing new performance baselines for financial question answering.
\item We demonstrate significant performance improvements through domain-specific optimizations, with retrieval accuracy increasing by 38.8\% and generation accuracy by 34.4\%.
\end{enumerate}

Our research seeks to answer the overarching question: 
\begin{quote}
Can the performance of Retrieval-Augmented Generation (RAG) systems for financial question answering be improved through (i) enhanced retrieval techniques, specifically changing the embedding model, applying chunk enrichment, and incorporating reranking, and (ii) improved generation techniques, specifically employing a reasoning-capable model?
\end{quote}


\section{Related Work}

\subsection{Existing Financial QA Datasets}
Specialized benchmarks for financial question answering (QA) have significantly advanced research in this domain. However, existing datasets each capture only a fragment of the broader landscape. Our benchmark is designed to complement these resources through practitioner-driven construction that reflects real financial workflows, while addressing gaps in regulatory coverage, document diversity, and cross-institutional analysis. This section reviews the principal financial QA datasets and situates our contribution within the broader benchmarking ecosystem.

\textbf{FinDER}~\cite{choi2025finder} marks a substantial step forward in financial QA benchmarking, providing 5,703 expert-annotated queries based on real-world financial search tasks. Designed to emulate the ambiguous and abbreviation-rich queries of finance professionals, FinDER challenges models to retrieve relevant information from large document collections. However, its scope is limited to U.S.~public companies’ annual reports. This constraint restricts generalization to international or multi-regulatory banking contexts.

\textbf{FinanceBench}~\cite{islam2023financebench}, comprises approximately 10,000 questions collected from filings of 40 U.S. public companies spanning the period 2015–2023. Each question is paired with an answer and the supporting context extracted from financial reports such as 10-K and 10-Q filings. Although comprehensive for single-company analysis, FinanceBench was not conceived for comparative studies across institutions—a critical capability for retrieval-augmented generation (RAG) systems employed in financial benchmarking and regulatory oversight. Furthermore, only a subset of 150 examples is publicly available, constraining its accessibility for open research.

Complementing these efforts, the \textbf{RAG Benchmark (Apple-10K-2022)}~\cite{lighthouz2024apple10k} provides a focused dataset constructed from Apple’s 10-K SEC filings for 2022. It consists of 100 query–response pairs with accompanying contextual information, serving as a targeted testbed for evaluating RAG models on single-institution financial disclosures.

Two additional datasets, \textbf{TAT-QA}~\cite{zhu2021tatqa_full} and \textbf{FinQA}~\cite{chen2021finqa}, extend the focus toward hybrid and numerical reasoning. TAT-QA contains approximately 2,800 hybrid contexts combining semi-structured tables and textual paragraphs, associated with roughly 16,500 financial questions. This design enables the evaluation of models that integrate textual and tabular data in complex reasoning tasks. FinQA, in turn, comprises around 2,800 financial reports yielding approximately 8,000 question–answer pairs, specifically targeting numerical reasoning over both structured and unstructured texts. Together, these datasets have been pivotal in advancing research on hybrid financial reasoning.

Building upon FinQA, \textbf{ConvFinQA}~\cite{chen2022convfinqa} introduces a conversational dimension to financial QA. It consists of 3,892 dialogues encompassing 14,115 questions, drawn from 10-K and 10-Q company reports. The dataset includes both simple conversations, generated by decomposing single multi-hop questions, and hybrid conversations, obtained by merging multiple reasoning chains. Each interaction was crafted by annotators with financial expertise, thereby enhancing realism and supporting the study of multi-turn reasoning in financial contexts.

Closer to our setting are two resources that extend this line of work along 
dimensions of length and retrieval realism. \textbf{DocFinQA}~\cite{reddy-etal-2024-docfinqa} pairs 7,437 
FinQA questions with their full SEC source reports, raising the average context 
length from under 700 words to approximately 123k words; the authors report that 
both retrieval-based pipelines and long-context models degrade markedly on the 
longest documents. \textbf{T\textsuperscript{2}-RAGBench}~\cite{strich-etal-2026-t2} 
instead addresses a structural property shared by FinQA, ConvFinQA and TAT-DQA: 
these datasets were constructed in an oracle-context setting, in which the 
relevant passage is supplied together with the question. Their questions are 
therefore context-dependent---the same question may admit different correct 
answers depending on which document is provided---which makes them unsuitable 
for evaluating retrieval. T\textsuperscript{2}-RAGBench recasts 23,088 
question--context--answer triples drawn from these sources into 
context-independent formulations over 7,318 documents. Both resources, however, 
remain confined to U.S.~filings and to single-institution analysis.

Finally, \textbf{HC3 Finance}~\cite{guo2023hc3}, a specialized subset of the broader HC3 dataset, evaluates model performance through direct comparison between human-generated answers and those produced by ChatGPT. This resource provides valuable insights into the relative strengths and weaknesses of large language models when confronted with domain-specific financial questions.
Table~\ref{tab:financial_datasets} provides a comprehensive comparison of FinRAG-QA  (our dataset) with existing financial and reasoning-oriented benchmarks, highlighting the unique characteristics and contributions of each resource.

\begin{table}[hbt!]
\centering
\caption{Comparison of financial and reasoning-oriented datasets used in retrieval and question answering tasks.}
\label{tab:financial_datasets}
\resizebox{\textwidth}{!}{%
\begin{tabular}{lp{0.45\textwidth}ccc}
\toprule
\textbf{Dataset} & \textbf{Description} & \textbf{\makecell{Publication \\ year}} & \textbf{\makecell{Number of \\ elements}} & \textbf{\makecell{Open-\\source}} \\
\midrule
FinRAG-QA (Ours) & \makecell[l]{The dataset consists of 999 rows \\ concerning financial indicators \\ (2019--2023) with ground-truth values.} & 2026 & \makecell{999 lines, \\ 209 documents} & Y \\
\midrule
FinDER & \makecell[l]{Expert-annotated dataset for \\
retrieval-augmented 
generation in \\ finance: query + 
evidence + answer \\ triplets 
drawn from real-world 
financial \\ queries with domain 
jargon, abbreviations, \\ and
ambiguous search behavior.} & 2025 & 5{,}703 & \makecell{Y} \\
\midrule
\makecell[l]{RAG benchmark \\ (Apple-10K-2022)} & \makecell[l]{Benchmark for Retrieval-Augmented \\ Generation in finance: 100 query-- \\ response--context triples from \\ Apple's 2022 10-K filing.} & 2024 & 100 & Y \\
\midrule
FinanceBench & \makecell[l]{Open-book financial QA benchmark: \\ 10{,}231 questions about publicly \\ traded companies with answers \\ and supporting evidence.} & 2023 & 10{,}231 & \makecell{N (only 150 \\ examples \\ public)} \\
\midrule
TAT-QA & \makecell[l]{Tabular And Textual QA over \\ hybrid contexts (text + tables) \\ in financial reports.} & 2021 & \makecell{16{,}552 \\ questions \\ across 2{,}757 \\ hybrid contexts} & Y \\
\midrule
FinQA & \makecell[l]{Numerical reasoning over financial \\ reports combining structured tables \\ and unstructured text with \\ annotated reasoning programs.} & 2021 & \makecell{8{,}281 QA \\ pairs across \\ 2{,}789 financial \\ reports} & Y \\
\midrule
ConvFinQA & \makecell[l]{Conversational financial QA with \\ multi-turn dialogues requiring \\ numerical reasoning over \\ company reports.} & 2022 & \makecell{3{,}037 train / \\ 421 dev / \\ 434 test \\ (conversation \\ level) \\ 11{,}104 train / \\ 1{,}490 dev / \\ 1{,}521 test \\ (turn level)} & Y \\
\midrule
DocFinQA & \makecell[l]{FinQA questions paired with full \\ SEC source reports; average \\ context $\approx$123k words.} & 2024 & \makecell{7{,}437 QA \\ pairs} & Y \\
\midrule
\makecell[l]{T\textsuperscript{2}-RAGBench} & \makecell[l]{Context-independent recast of \\ FinQA, ConvFinQA and TAT-DQA \\ for retrieval evaluation over \\ text-and-table documents.} & 2026 & \makecell{23{,}088 triples \\ across 7{,}318 \\ documents} & Y \\
\midrule
\makecell[l]{HC3 Finance} & \makecell[l]{Human--ChatGPT comparison corpus \\ in finance with paired human and \\ ChatGPT answers to finance-related \\ questions.} & 2023 & 48{,}644 & Y \\
\bottomrule
\end{tabular}%
}
\end{table}

\subsection{Dataset’s Contribution}
Existing financial QA benchmarks show gaps: they largely center on U.S. regulation, overlook European banking standards such as Basel III Pillar 3 disclosures, and favor single-institution analysis over cross-bank comparisons, limiting their usefulness for analysts and regulators. Many also reflect academic priorities more than professional workflows, weakening ecological validity. Our dataset addresses these limitations by integrating 89 Pillar 3 reports and 120 annual reports, an international sample of 24 major banks across Europe and the United States, 
enabling systematic, multi-institutional benchmarking of RAG systems on regulatory compliance documents.
These documents are longer, on average, than 
those of any prior financial QA benchmark.
Designed in collaboration with financial professionals, it targets ten standardized indicators (e.g., CET1 ratio, asset classes).

Rather than replacing existing datasets, our benchmark complements prior work by filling a specific and critical gap. It enables:
\begin{itemize}
    \item Cross-institutional analysis for competitive and regulatory benchmarking,
    \item Regulatory compliance evaluation within the banking context,
    \item Systematic and reproducible metric extraction for professional workflows.
\end{itemize}

This unique positioning establishes our benchmark as an essential addition to the research ecosystem for robust and comprehensive RAG evaluation in financial question answering applications.

\section{Dataset}
Our benchmark dataset\footnote{Compiled by Intesa Sanpaolo collecting public data from end of year financial and regulatory disclosures.} contains 999 rows representing financial questions across 10 key financial indicators.

\subsection{Dataset Structure}
The dataset structure includes the following components (see also \ref{tab:dataset}): 
\begin{itemize}
  \item \textbf{Data}: the name of the indicator of interest (10 data values). The financial indicators covered include:
 Common Equity Tier 1 (CET1) capital, Day One Profit, Total Additional Valuation Adjustments (AVA), Total Assets, Total Fair Value Level 1, 2, and 3 Assets, Total Fair Value Level 1, 2, and 3 Liabilities.

  \item \textbf{Year}: the year to which the data refer, ranging from 2019 to 2023 (inclusive) (5 years). 
  \item \textbf{Bank}: the bank to which the data refer (24 major international banks).

  \item \textbf{Doc Type}: the type of document in which the data appears. It can be a \textit{Financial Report} or a \textit{Pillar 3} document.

  \item \textbf{Query}: the queries submitted to the model. These follow a standardised format:
  \begin{quote}
  \iq{What is the consolidated \textless{}Data\textgreater{} value in millions for \textless{}Bank\textgreater{} for the year \textless{}Year\textgreater{}?}
  \end{quote}
 
  \item \textbf{Value}: indicator's value, which serves as ground truth and is a decimal number in millions, in the currency of the source document. All ground truths are \iq{simple} values, i.e., they are not derived from a combination of other values.

  \item \textbf{Numerical}: indicates whether the value is present in clear numerical form. For example, sometimes there are indicators whose value is not available (21 cases). These can be indicated in natural language or expressed in tables with dashes ($-$). The reason this column was added is to better manage the retrieval phase.

  \item \textbf{In table}: indicates whether the value appears, in the source document, in a table (Y), or not (N).

  \item \textbf{Source Document Year}: specifies the publication year of the source document. Financial reports commonly include comparative data, presenting figures for the current reporting period alongside those from one or more previous years. Consequently, the value for a specific financial indicator and year (e.g., 2022) might be sourced from a document published in a subsequent year (e.g., the 2023 annual report). This means the \textbf{Source Document Year} does not necessarily match the \textbf{Year} of the data point, a characteristic that reflects real-world financial analysis practices and introduces a temporal challenge for retrieval systems.
  \end{itemize}
 
\begin{table}[hbt!]
    \centering
    \resizebox{\textwidth}{!}{%
    \begin{tabular}{|c|c|c|c|c|c|c|c|c|}
        \hline
        \textbf{Data} & \textbf{Year} & \textbf{Bank} & \textbf{\begin{tabular}[c]{@{}c@{}}Doc \\ Type\end{tabular}} & \textbf{Query} & \textbf{Value} & \textbf{Numerical} & \textbf{\begin{tabular}[c]{@{}c@{}}In \\ table\end{tabular}} & \textbf{\begin{tabular}[c]{@{}c@{}}Source \\ Document \\ Year\end{tabular}} \\ \hline
        \begin{tabular}[c]{@{}c@{}}Total \\ Assets\end{tabular} & 2023 & \begin{tabular}[c]{@{}c@{}}Banco \\ BPM\end{tabular} & Report & \begin{tabular}[c]{@{}c@{}}What is the \\ Total Assets \\ value in million \\ for Banco BPM \\ for the year \\ 2023?\end{tabular} & 202131.973 & Y & Y & 2023 \\ \hline
    \end{tabular}%
    }
    \caption{Example of a row in the dataset}
    \label{tab:dataset}
\end{table}

\subsection{Dataset Statistics}
Table~\ref{tab:dataset_statistics} reports the corpus statistics.
\begin{table}[hbt!]
    \centering
    \begin{tabular}{|l|c|l|c|}
        \hline
        \textbf{\begin{tabular}[c]{@{}l@{}}Total number \\ of documents\end{tabular}} & 209 & \textbf{\begin{tabular}[c]{@{}l@{}}Total number \\ of chunks\end{tabular}} & 150437 \\ \hline
        \textbf{\begin{tabular}[c]{@{}l@{}}Mean number \\ of chunks\end{tabular}} & 720 & \textbf{\begin{tabular}[c]{@{}l@{}}Max number \\ of chunks\end{tabular}} & 2345 \\ \hline
        \textbf{\begin{tabular}[c]{@{}l@{}}Mean number \\ of words/chunk\end{tabular}} & 276 & \textbf{\begin{tabular}[c]{@{}l@{}}Max number \\ of words/chunk\end{tabular}} & 19625 \\ \hline
        \textbf{\begin{tabular}[c]{@{}l@{}}Mean number \\ of words/document\end{tabular}} & 198515 & \textbf{\begin{tabular}[c]{@{}l@{}}Max number \\ of words/document\end{tabular}} & 556505 \\ \hline
    \end{tabular}
    \caption{Detailed statistics of the dataset.}
    \label{tab:dataset_statistics}
\end{table}


The corpus of 209 financial documents comprises:
\begin{itemize}
\item 120 Annual Financial Reports
\item 89 Pillar 3 Reports
\end{itemize}

These documents span five years (2019-2023) and cover an international sample of 24 major international banks from Europe (e.g. Intesa Sanpaolo, UniCredit, BNP Paribas, Deutsche Bank, etc.) and USA (JPMorgan Chase, Bank of America, Citigroup, Wells Fargo).

\subsection{Dataset Characteristics}
The dataset is characterized by several inherent challenges that mirror real-world financial analysis. These include the structural complexity of financial documents, which often span hundreds of pages with intricate hierarchies; the prevalence of technical jargon and domain-specific abbreviations; the necessity for precise numerical reasoning to extract and interpret quantitative data, including unit conversion to match the ground truth required in millions; and the inhomogeneous presentation formats, with information appearing in tables, embedded within text, or occasionally absent.

\section{Experimental Setup}
This section describes the architecture and evaluation methodology adopted to assess the effectiveness of the proposed RAG pipeline for financial question answering, detailing each stage from document ingestion to generation.
All experiments were conducted between late 2024 and early 2025, using the most advanced generation models available at that time.

\subsection{RAG Architecture}
Retrieval-Augmented Generation (RAG)~\cite{NEURIPS2020_6b493230} is an architectural paradigm designed to enhance the accuracy and reliability of Large Language Models (LLMs) by grounding their responses in external knowledge bases. Instead of relying solely on its internal, pre-trained knowledge, a RAG system first retrieves relevant information from a specified corpus of documents and then uses this information to generate a more informed and contextually accurate answer.
Our RAG architecture can be deconstructed into a multi-stage pipeline: document ingestion and preprocessing, chunk indexing, retrieval, reranking and generation.

\subsubsection{Document Ingestion and Preprocessing}

PDF documents were processed using Microsoft Azure Document Intelligence~\cite{azure2024docintelligence} to extract text in Markdown format. We developed a modified version of LangChain's MarkdownHeaderTextSplitter to segment documents based on hierarchical headings, maximizing granularity to improve retrieval accuracy. The preprocessing pipeline includes OCR-based text extraction with layout preservation, Markdown conversion maintaining document structure, hierarchical chunking based on heading levels, page range tracking for each chunk, and metadata preservation (bank name, year, document type).

\subsubsection{Chunk Indexing}
The processed chunks are transformed into numerical representations (embeddings) and stored in a specialized vector database for efficient searching.
We evaluated two leading dense embedding models to understand their impact on retrieval performance within the financial domain. All chunks were indexed in a Qdrant~\cite{qdrant2024vectordb} vector store. The selected models were:
\begin{itemize}
    \item \textbf{OpenAI text-embedding-3-large (baseline)~\cite{openai2024text_embedding_3_large}}: a powerful, general-purpose model with 3072 dimensions. It serves as our baseline due to its strong performance on broad benchmarks like MTEB - Massive Text Embedding Benchmark (64.6\%)~\cite{openai2025_obtaining_embeddings}and its widespread adoption in production RAG systems.
    \item \textbf{VoyageAI voyage-3-large~\cite{voyageai2025_voyage_3_large}}: a state-of-the-art model specifically optimized for retrieval tasks. It has demonstrated superior performance over other models, including a 9.74\% higher NDCG@10 than \textit{text-embedding-3-large} across diverse datasets~\cite{voyageai2025_voyage_3_large}.
\end{itemize}

\subsubsection{Contextual Chunk Enrichment}
Traditional RAG systems split documents into small chunks for retrieval and semantic search. However, this splitting often discards important context, making individual chunks hard to understand and leading to weaker retrieval results. 

To solve this issue, we apply the Contextual Retrieval technique inspired by Anthropic~\cite{anthropic2024contextual}. For each chunk, we automatically generate a concise, synthetic contextual summary (typically 50–100 tokens) using GPT-4.1 that has a 1M token context window. This summary describes how the chunk fits within the overall document, in order that even small document fragments retain enough background information to be accurately retrieved and interpreted. The context is then prepended to the original chunk before embedding or indexing, supplying retrieval models with rich, domain-aware representations.

The developed enrichment process uses document structure and heading hierarchies to identify the right context for each chunk, even in very large files that exceed the model’s context window. By dynamically adjusting parameters (such as the allowed difference in heading levels), we ensure each chunk’s context remains accurate while staying within computational and token limits.


\subsubsection{Reranking}
Initial retrieval, based on semantic search, is optimized for speed over a large corpus but may return chunks that are only broadly relevant. Reranking introduces a second, more precise filtering stage to refine these initial results. It employs a more powerful but computationally intensive model, which re-evaluates the top-k relevant chunks by jointly considering the query and each chunk's content. This deep semantic analysis provides a more accurate relevance score than the initial vector similarity search.

The model used in this work is Cohere rerank-3.5~\cite{cohere2025rerank35}, a cross-encoder trained on large-scale data including web search, question-answering, and multilingual corpora. The reranking process initially retrieves 100 chunks, then narrows down to the top-k through relevance scoring.


\subsubsection{Generation Models}
 In the final phase, the top-ranked, context-rich chunks are combined with the original user query and passed as context to a generator LLM. The LLM then synthesizes this information to produce the final answer.
Given that experiments were carried out in late 2024 and early 2025, we evaluate two generation models available at that time:

\begin{itemize}
    \item \textbf{GPT-4o}: a highly capable, general-purpose multimodal model used as a baseline. It features a 128,000-token context window and is known for its strong performance across a wide range of tasks.
    \item \textbf{GPT-o1-high}: a model specifically optimized for complex reasoning tasks. It provides a 200,000-token context window and is configured to use its maximum reasoning effort, prioritizing accuracy over response latency. This model is selected to test the hypothesis that enhanced reasoning capabilities improve performance on numerical extraction and interpretation tasks.
\end{itemize}

\subsection{Evaluation Metrics}

\subsubsection{Retrieval Phase}
We evaluate retrieval quality using the Normalized Discounted Cumulative Gain at k (NDCG@k) metric~\cite{jarvelin2002cumulated, evidentlyai2025ndcg}, which assesses the quality of a ranked list by prioritizing relevant items in top positions. We test for k values of 1, 10, and 20. NDCG@k is defined as:
\begin{equation}
NDCG@k = \frac{DCG@k}{IDCG@k} \ ,
\end{equation}
where DCG@k (Discounted Cumulative Gain) is calculated as:
\begin{equation}
DCG@k = \sum_{i=1}^{k} \frac{G_i}{\log_2(i + 1)} \ .
\end{equation}
Here, $G_i$ is the relevance score of the chunk at rank $i$, and $IDCG@k$ is the DCG of an ideal ranking.

Since our dataset does not explicitly label "golden chunks," we dynamically compute the relevance score $G_i$ for each retrieved chunk. A dedicated function searches for the numerical ground truth value within the chunk's text, accounting for various formatting permutations (e.g., different decimal and thousand separators). The relevance score $G_i$ is set to the number of times the ground truth value is found within the chunk; chunks not containing the value receive a score of 0. Queries for which the ground truth is not a numerical value are excluded from this evaluation.

The result is a value between 0 and 1, where 1 represents a perfect ranking
(all relevant items are in the highest positions) and 0 represents the absence of
relevance in the first k results. In general, the higher the NDCG, the better
the quality of the ranking compared to the ground truth.

\subsubsection{Generation Phase}
To isolate the generator's performance from retrieval errors, we evaluate generation accuracy only on queries where the ground truth value is present within the top-k relevant chunks. The performance is measured using accuracy, defined as the proportion of correct answers. An answer is considered correct if the generated value matches the ground truth value, if both are null, or if the absolute difference between the numerical values is less than or equal to 0.01. The accuracy is calculated as:
\begin{equation}
\text{Accuracy} = \frac{1}{N} \sum_{i=1}^{N} \mathbb{I}(\hat{y}_i \approx y_i)
\end{equation}
where $N$ is the number of evaluated queries, $\hat{y}_i$ is the predicted value, $y_i$ is the ground truth, and $\mathbb{I}(\cdot)$ is the indicator function that is 1 if the condition is met and 0 otherwise.

\section{Results}
Our experimental evaluation is structured in two phases: retrieval and generation. We assess various configurations to identify the optimal components for a financial RAG pipeline.

\subsection{Retrieval Performance}
Table \ref{tab:retrieval_results} summarizes the performance of different embedding and retrieval strategies. All values were calculated from a total of 978 queries (999 total minus 21 where the ground truth is not numerical).

\begin{table}[hbt!]
\centering
\caption{Retrieval results (NDCG@k) for different scenarios. The numbers in parentheses indicate the count of queries for which the ground truth was found in the top-k retrieved chunks. Bold values indicate the best performance for each k.}
\label{tab:retrieval_results}
\resizebox{\textwidth}{!}{%
\begin{tabular}{lcccc}
\toprule
\textbf{Scenario} & \textbf{NDCG@1} & \textbf{NDCG@10} & \textbf{NDCG@20} & \textbf{\makecell{Avg. \\ Time (s)}} \\
\midrule
OpenAI embeddings (baseline) & 0.163 (159) & 0.322 (524) & 0.347 (649) & 1.9 \\
OpenAI embeddings + headings & 0.166 (162) & 0.315 (508) & 0.346 (651) & 1.4 \\
OpenAI contextualized embeddings & 0.248 (242) & 0.459 (723) & 0.478 (830) & 1.8 \\
OpenAI contextualized + reranking & 0.332 (324) & 0.494 (698) & 0.511 (795) & 4.3 \\
\textbf{VoyageAI contextualized embeddings} & \textbf{0.510} (498) & \textbf{0.710} (929) & \textbf{0.705} (956) & \textbf{1.3} \\
VoyageAI contextualized + reranking & 0.342 (334) & 0.498 (700) & 0.512 (791) & 3.4 \\
\bottomrule
\end{tabular}%
}
\end{table}

The key findings from the retrieval evaluation are:
\begin{itemize}
    \item \textbf{Contextual Enrichment and Advanced Embeddings}: the combination of contextual chunk enrichment and the VoyageAI embedding model yields a dramatic performance increase. This configuration achieved a peak NDCG@10 of 0.710, a 38.8\% absolute improvement over the 0.322 NDCG@10 of the OpenAI baseline. This configuration also had the fastest average retrieval time.
    \item \textbf{Reranking Effects}: reranking produced mixed results. It improved the performance of the OpenAI contextualized embeddings scenario by an average of 5.1\% across k values. However, it degraded the performance of the superior VoyageAI configuration by an average of 19.1\%, suggesting that reranking may be detrimental when the initial retrieval quality is already high.
    \item \textbf{Impact of k}: increasing k from 10 to 20 for the best-performing scenario (VoyageAI contextualized) slightly decreased the NDCG score (from 0.710 to 0.705) but increased the number of queries with a retrieved ground truth from 929 to 956 (out of 978). This highlights a trade-off between ranking quality and overall recall.
    \item \textbf{Structural vs. Semantic Enrichment}: using document headings for enrichment provided negligible benefit over the baseline, indicating that rich semantic context is more critical for retrieval than structural metadata alone.
\end{itemize}

\subsection{Generation Performance}
Generation performance was evaluated using accuracy, defined as the proportion of generated answers matching the ground truth. To isolate the generator's capability, we only evaluated queries where the ground truth was present in the retrieved context. Based on retrieval results, reranking scenarios were excluded from this phase.

Table \ref{tab:generation_results} presents the accuracy for the GPT-4o (baseline) and GPT-o1-high (reasoning-optimized) models across different retrieval scenarios and k values.

\begin{table}[hbt!]
\centering
\caption{Generation accuracy for different models and retrieval scenarios. The number in parentheses indicates the count of queries evaluated.}
\label{tab:generation_results}
\begin{tabular}{llccc}
\toprule
\textbf{Scenario} & \textbf{Model} & \textbf{k=1} & \textbf{k=10} & \textbf{k=20} \\
\midrule
\multirow{2}{*}{OpenAI embeddings (baseline)} & 4o & 0.306 (180) & 0.402 (545) & 0.399 (670) \\
& o1-high & 0.544 (180) & 0.747 (545) & 0.755 (670) \\
\midrule
\multirow{2}{*}{\begin{tabular}[l]{@{}l@{}}OpenAI embeddings +\\headings, prompt + headings\end{tabular}} & 4o & 0.399 (183) & 0.454 (529) & 0.406 (672) \\
& o1-high & 0.623 (183) & 0.798 (529) & 0.808 (672) \\
\midrule
\multirow{2}{*}{OpenAI contextualized embeddings} & 4o & 0.490 (263) & 0.504 (744) & 0.497 (851) \\
& o1-high & 0.722 (263) & 0.823 (744) & 0.812 (851) \\
\midrule
\multirow{2}{*}{VoyageAI contextualized embeddings} & 4o & 0.455 (519) & 0.467 (950) & 0.436 (977) \\
& o1-high & 0.852 (519) & 0.822 (950) & 0.811 (977) \\
\bottomrule
\end{tabular}
\end{table}

The main insights from the generation phase are:
\begin{itemize}
    \item \textbf{Reasoning Model Superiority}: the reasoning-optimized GPT-o1-high model consistently and significantly outperformed the GPT-4o baseline across all scenarios. The weighted average accuracy for GPT-o1-high was 79.0\%, compared to 44.6\% for GPT-4o, representing a 34.4\% absolute improvement.
    \item \textbf{Optimal Context Size (k)}: for generation, providing a smaller, more focused context (k=1) often yielded the highest accuracy, particularly for the top-performing GPT-o1-high model. This suggests that including more, potentially noisy, chunks can degrade the generator's ability to pinpoint the correct answer, even if the ground truth is present.
    \item \textbf{Performance vs. Latency}: while GPT-o1-high delivered superior accuracy, its average generation time was approximately 20 times longer than GPT-4o (35.0s vs. 1.8s). This highlights a critical trade-off between accuracy and response latency for practical applications.
\end{itemize}

\section{Conclusions}

\subsection{Key findings}
This work presents a comprehensive evaluation of RAG systems for financial question answering, introducing a novel benchmark dataset and demonstrating significant performance improvements through optimizations. Our key contributions include:

\begin{enumerate}
\item A novel benchmark dataset with 999 expert-validated questions across 10 financial indicators and 24 major banks, providing a rigorous testbed for financial QA systems.

\item Substantial performance improvements through contextual chunk enrichment and advanced embedding models, with retrieval accuracy increasing by 38.8\% (NDCG@10: 32.2\% to 71.0\%).

\item Demonstration of reasoning model superiority, with GPT-o1-high achieving 34.4\% higher generation accuracy compared to GPT-4o (79.0\% vs 44.6\%).

\item Comprehensive evaluation of multiple RAG configurations, revealing that domain-specific optimizations significantly outperform generic approaches.
\end{enumerate}

Our findings highlight several important insights for financial RAG system development. Contextual enrichment of document chunks proves more valuable than structural enhancements, suggesting that semantic understanding trumps syntactic organization. Advanced embedding models specifically trained for financial domains demonstrate substantial improvements over general-purpose alternatives. Reasoning-optimized language models show particular strength in tasks requiring precise numerical extraction and interpretation, which are common in financial analysis.

Our work establishes a foundation for advanced financial document analysis systems, and the benchmark released alongside it is a contribution of lasting value: while specific architectural choices will continue to evolve, a curated, expert-validated dataset spanning European and U.S. regulatory disclosures remains a stable standard of comparison for measuring that progress. We therefore expect the benchmark to serve not only as the evaluation vehicle for the present study, but as shared infrastructure for the financial NLP community.

\subsection{Limitations}

The following limitations affect our current approach:

\begin{itemize}
\item \textbf{Computational Cost}: contextual enrichment requires processing entire documents with large language models, introducing significant computational overhead that may be prohibitive for smaller organizations.

\item \textbf{Reranking Inconsistency}: while reranking typically improves retrieval performance, our results show mixed outcomes, suggesting that the interaction between initial retrieval quality and reranking effectiveness requires further investigation.

\item \textbf{Reproducibility Concerns}: the stochastic nature of reasoning models, particularly GPT-o1-high, can affect result reproducibility, which is critical for financial applications requiring audit trails.

\item \textbf{Temporal Validity}: the experiments were conducted in 2024–2025 using GPT-4o and GPT-o1-high, the OpenAI models available at that time. Subsequent generations of language models may yield different accuracy and latency trade-offs, and the reported results should be interpreted as a snapshot of performance achievable with 2024–2025 technology.

\end{itemize}

%

\subsection{Future Work}
While the proposed RAG system already provides tangible support for analysts' workflows, the residual error rate remains too high. Future work should target a systematic reduction of this margin, according to the following directions: 

\begin{itemize}
\item \textbf{Efficiency Optimization}: developing more efficient contextual enrichment methods, potentially using smaller specialized models for context generation or selective enrichment based on chunk importance.

\item \textbf{Refined Preprocessing}: enhancing document preprocessing by using cost-effective models to pre-filter chunks, selecting only those containing tables or relevant financial data to improve the efficiency of subsequent pipeline stages.

\item \textbf{Alternative Reranking Strategies}: exploring alternative reranking methods, such as using more economical proprietary models or fine-tuning open-weight models, to address the inconsistent performance and high cost of current rerankers.

\item \textbf{Advanced Generation Models}: experimenting with newer and more advanced language models in the generation phase to evaluate potential improvements in accuracy and reasoning capabilities.

\end{itemize}

\printbibliography

\end{document}